# Explore the Knowledge contained in Network Weights to Obtain Sparse Neural Networks

Mengqiao Han, Xiabi Liu, Zhaoyang Hai, Zhengwen Li

## Abstract

*Sparse neural networks are important for achieving better generalization and enhancing computation efficiency. This paper proposes a novel learning approach to obtain sparse fully connected layers in neural networks (NNs) automatically. We design a switcher neural network (SNN) to optimize the structure of the task neural network (TNN). The SNN takes the weights of the TNN as the inputs and its outputs are used to switch the connections of TNN. In this way, the knowledge contained in the weights of TNN is explored to determine the importance of each connection and the structure of TNN consequently. The SNN and TNN are learned alternately with stochastic gradient descent (SGD) optimization, targeting at a common objective. After learning, we achieve the optimal structure and the optimal parameters of the TNN simultaneously. In order to evaluate the proposed approach, we conduct image classification experiments on various network structures and datasets. The network structures include LeNet, ResNet18, ResNet34, VggNet16 and MobileNet. The datasets include MNIST, CIFAR10 and CIFAR100. The experimental results show that our approach can stably lead to sparse and well-performing fully connected layers in NNs.*

## 1. Introduction

Deep neural network (DNN) as a flexible function approximator has been very successful in a broad range of tasks [1, 2]. However, previous researches have shown that they are over-parameterized because some weights can be pruned without reducing accuracy [3, 4, 5]. Furthermore, they can easily overfit and even memorize random patterns in the data, if not properly regularized [6].

A way to address above issues is to reduce the complexity of DNNs. There have been studies on this topic, including network pruning [7, 8], knowledge distillation [9, 10], network architecture search [11], introduce sparse regularization [12, 13], and adaptive model compression [14]. The existing methods usually sacrifice the network performance for sparser structures. Furthermore, additional fine-tuning process is required in some existing methods.

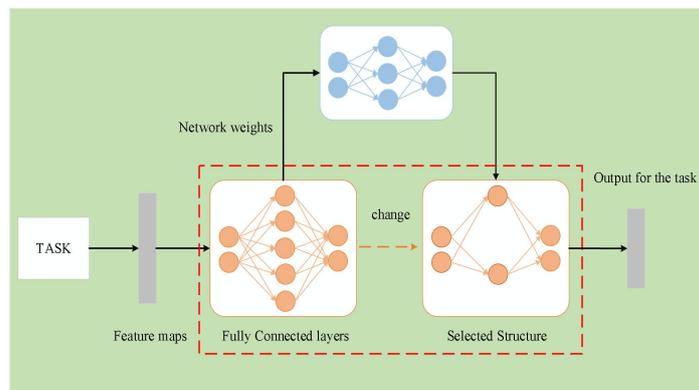

Figure 1. An overview of the proposed framework.

In the mammalian brain, synapses are created in the first few months of a child's development, followed by gradual pruning of little-used connections, falling to typical adult values [15, 16]. It means that the cognitive process of the mammalian brain is to dynamically pruning synaptic connections and renew the remaining ones. This inspires our work in this paper. We imagine there is an additional schema of simplifying the structure of the network, along with the optimization of its weights. We introduce a switcher neural network (SNN) to perform this task, which utilizes the knowledge contained in the weights of DNN to switch each connection in DNN and simplify its structure automatically. Based on SNN, we propose an efficient and novel learning framework, which is shown in Figure 1. We can see that our framework is composed of two parts, task neural network (TNN) and SNN. TNN is the usual network, for finding a winning-ticket [8]. The SNN explores the knowledge contained in the weights of TNN to optimize its structure. Then, the TNN renews it weights to adapt to this new structure. This leads to an alternating optimization of SNN and TNN. It can be seen as gradually pruning unimportant connections during the training and optimizing the remaining connections. To achieve the good results of sparse learning, the SNN structure is designed to be U-net like, composed of an encoder and a decoder. The proposed SNN-TNN framework let us obtain the optimal structure and the optimal parameters of TNN simultaneously and directly, without retraining or fine-tuning. Furthermore, the loss functions used in the two procedures of alternate learning are both the performance evaluation of TNN, such as commonly used Cross Entropy (CE) in image classification applications in this paper. So, we avoid imposing unreasonable sparsity on over-parameterized models.

Some existing methods for simplifying networks also rely on the knowledge contained in network weights, but usually simple and setting-by-experience, such as low weights indicating unimportant connections [7]. Differently, we try to represent and mine more sophisticated knowledge in network weights with a DNN, i.e., SNN. To our best knowledge, this is the first time of such efforts.

We apply the proposed approach to obtain sparse fully connected layers (FCL) in DNNs. The FCLs, even if they are in the minority in number, are responsible for the majority of the parameters in mainly used DNNs [17]. Experiments on various types of neural network models and datasets are conducted to confirm the effectiveness of our method.

The main contributions of this paper are summarized as follows:

1. We propose a novel network structure learning approach by introducing the idea of switcher neural network (SNN) to represent and explore the knowledge contained in the network weights for switching connections in the network.
2. The structure of SNN is designed based on the thinking of fully exploring the relationship among weights to determine the importance of each connection. It is divided into an encoder and a decoder, like U-net shape, But the feature maps generated by the encoder need to be transformed before they are inserted into the decoder path.
3. We develop a learning algorithm to optimize the SNN and the TNN alternately under the same learning objective. It can learn structure and parameters at the same time, without additional regular items, re-training and fine-tuning.
4. The proposed approach is applied to optimize the FCLs of DDNs for classification. It leads to not only sparser but also more accurate classifiers in the experiments on various types of networks and various datasets.

## 2. Related Work

The main methods on the topic of sparse DNNs are introduced briefly as follows.

**Pruning.** Han et al. [7] showed per-weight magnitude-based pruning substantially reduces the size of image-recognition networks. Han et al. [18] and Jin et al. [19] restored pruned connections to increase network capacity after small weights have been pruned and surviving weights fine-tuned. Other proposed pruning heuristics include redundancy [20] and energy efficiency [21]. Frankle et al. [22] demonstrates that dense feed-forward neural networks contain winning-tickets, which are sparse sub-networks, when reset to their initialization and trained separately, without accuracy drop.

**Regularization.** Srinivas et al. [23] learned gating variables that minimize the number of nonzero parameters. Wang et al. [12] used the characteristics of convolutional layers to reduce the memory and computational cost of FCLs. Louizos et al. [24] smoothed the expected L0 regularized objective with continuous distributions to achieve the L0 norm regularization of parameters. Wen et al. [25] applied group LASSO to regularize multiple structures in DNNs. Zhu et al. [26] investigated that DNNs trained by group LASSO constraint still exist substantial filter correlation among the convolution filters, and suggested to suppress such correlation with a new kind of constraint called decorrelation regularization.

**Knowledge distillation and Dictionary learning.** Distillation [9] trains small networks to mimic the behavior of large networks. Yang et al. [27] proposed a training algorithm based on the concept of sparse representation, which consists of structure optimization and weight update.

**Structured dropout probabilities and Bayesian model.** Molchanov et al. [28] extended variational dropout to the case when dropout rates are unbounded. Neklyudov et al. [39] injected noises to the neurons' outputs while keeping the weights unregularized, which leads to structured sparsity by removing elements with a low signal noise ratio from the computation graph. Louizos et al. [30] explicitly prune and sparsify networks during training as dropout probabilities for some weights reaching 1. Dai et al. [31] penalized the inter-layer mutual information using a variational approximation.

**Architecture selection.** Architecture search method is to explore the space of potential models automatically, but perform neural architecture search using reinforcement learning is expensive [32]. And some works usually achieve the lower performance compared to their backbone network [33] or require additional fine-tuning process [34]. Recent work, Ahn et al. [11] use selector to select the model with the highest probability from estimator to perform the task. Wortsman et al. [35] enable channels to form connections independent of each other, and adjust the update rule.

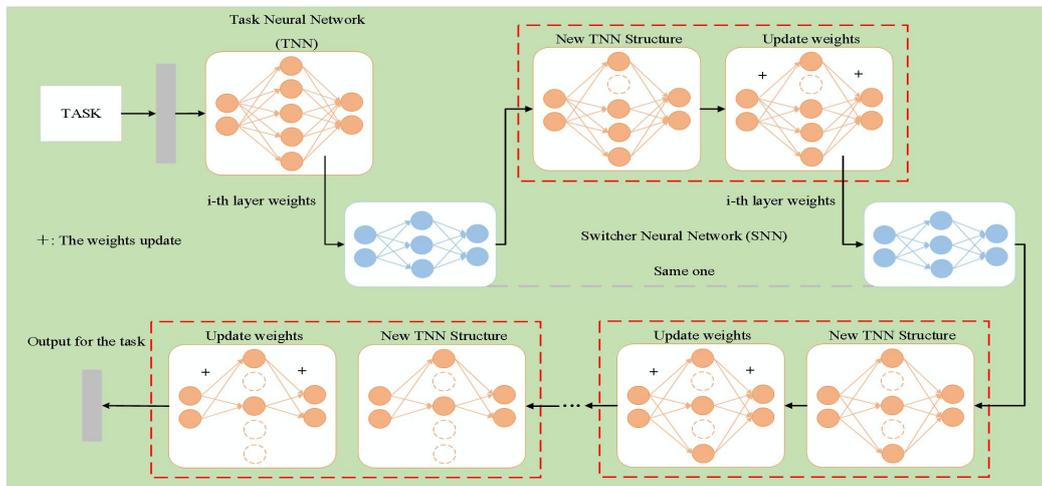

Figure 2. A graphical representation of the proposed framework, which consists of task neural network (TNN) and switcher

neural network (SNN). TNN is used for classification in this paper. SNN (discuss in section 3.2) learns the weight matrix of the TNN.

## 3. The Proposed Approach

The core idea of our proposed approach is to explore the knowledge contained in the network weights for determining the importance of each connection of the network. The SNN is introduced to fulfill this idea, which takes the weight matrix between the i-th layer and the i+1-th layer of the TNN as the inputs and outputs the values for switching the connections in the i+1-th layer of TNN.

### 3.1 SNN Guided TNN Optimization

#### 3.1.1 The Network Architecture

The overall framework of the proposed approach is illustrated in Figure 2, which consists of two components: the TNN is used for performing actual tasks, such as classification in the experiments of this paper, and the SNN represents and explores the knowledge contained in the weights of TNN for optimizing TNN structure. Actually, the outputs of SNN are multiplied with the weights of corresponding connections in TNN, thus it plays a role of weighting connections and the zero outputs of SNN means the pruning of corresponding connections. In this paper, we explore the weight matrix between the i-th layer and the i+1-th layer to switch connections in the i+1-th layer. We consider two strategies. The first strategy is to take all the output connections of a neuron as a whole. The processing effect is equal with switching each neuron in i-th layer. Thus the SNN outputs a different value for each neuron and let it multiplied with the output of the corresponding neuron. The second strategy is to single out each connection. Thus the SNN outputs a different value for each connection and let it multiplied with the weight of the corresponding connection.

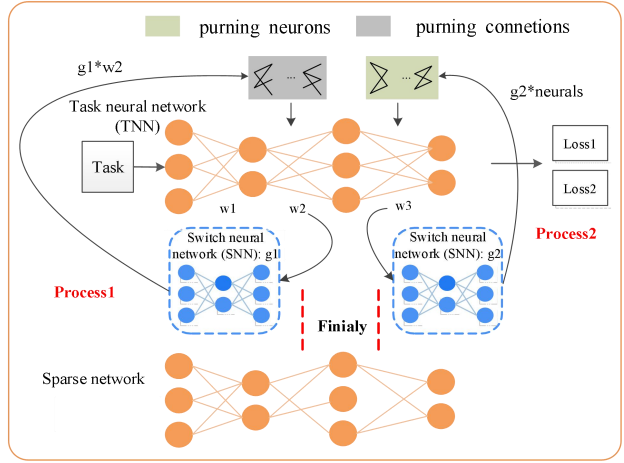

Figure 3. The example of SNN guided TNN Optimization for FCLs in DNN classifiers.

Now we explain the above SNN-TNN schema by taking the learning of sparse FCLs in DNN classifiers as an example. The resultant network architecture is illustrated in Figure 3, where Process1 and Process2 represent the process of connection switching and that of neuron switching, respectively. Let m and n be the number of neurons in the i-th and i+1-th layer, respectively, then in both process1 and process 2, the SNN will accept m*n weights. But it will output m*n values in Process1 to indicating the importance degree of each connection. Then these the matrix constructed by these m*n values are multiplied with the weight matrix, such as shown as g1*w2 in Figure 3. The SNN output m values in Process2 to indicating the importance degree of each neuron. These the vector constructed by these m values are multiplied with the input data of this layer, such as shown as neuron*g2 in Figure 3.

The following formulas help to better express our method. The forward computation of three-layer fully connected networks as TNN is

$$\hat{y} = W_3 \Phi(W_2 \Phi(W_1 X))$$

Where Wk (k=1-3) denotes the weight matrix of the k-th layer, $\Phi$ denotes the activation function, X denotes the input data, and $\hat{y}$ denotes the predicted value of the network.

We set SNNs for each FCL of the TNN. Let gk be a SNN for the k-th layer.

The output of TNN for our first strategy of switching neurons is

$$\hat{y} = \Phi((\Phi(W_1 X)) \Theta Z_2 W_2) \Theta Z_3 W_3$$

$$W_{finial} = W_{optimal}$$

Then the output of TNN for our second strategy of switching connections is

$$\hat{y} = W_3 \Theta Z_3 (\Phi(W_2 \Theta Z_2 (\Phi(W_1 X))))$$

$$W_{finial} = W_{optimal} \Theta g_k$$

where $\Theta$ denotes the element wise product.

### 3.1.2 Alternating Learning Algorithm

The satisfactory TNN is determined by the optimal parameters of SNN and the optimal parameters of TNN. In order to optimize the two networks, we present a learning algorithm to train these two parts together, which is composed of two phases that are performed alternatively and iteratively. In the first phase, the TNN is remained and the SNN parameters are optimized to choose a reasonable structure of TNN. In the second phase, the SNN is remained and only TNN parameters are optimized to adapt to this new structure. We use the same loss function in these two phases, such as the CE loss in the applications of this paper. By using this strategy, the network sparsity is not required explicitly, but is achieved indirectly by requiring the optimized performance in applications. It is similar to the working way of our mammalian brain that continuously and dynamically optimizes the synaptic connections in the process of learning new things [15, 16]. We believe that it is more natural for the network to find the optimal structure through targeting the direct application goals and exploring the relationship among its own neuron connection than the way of exerting human-experienced regularization terms. The regularization items on sparsity could impose unreasonable sparsity to cause accuracy drop. Our experimental results confirm that our learning strategy can effectively avoid this problem and leads to even better performance with sparser structures.

The above alternative learning algorithm is summarized in Algorithm.

> **Algorithm:** The alternative learning algorithm of TNN and SNN.

```
input: The datasets X. W_k is the weight matrix in k-the layer
       of the TNN, W^g is weight matrix in the SNN,
       non-linear transfer function Φ, the number of maximal
       iterations T.
       ~ denotes the optimized parameters.
       ^ denotes the fixed parameters.
       * denotes the optimal network parameters.
output: The TNN parameters $W_k$, and the SNN parameters
       $W^g$
initialize: Both W_k and W^g use Kaiming normal
       distribution
for epoch=0 to T do:

    if epoch%2 == 0:

        $^\wedge W_k$  and  $\sim W^g$

    elif epoch%2 != 0:

        $\sim W_k$  and  $^\wedge W^g$

end
```

## 3.2 SNN Structure

The purpose of SNN is to determine the importance of the connections based on the knowledge contained in TNN's weights. The corresponding structure of SNN is designed as that illustrated in Figure 4, which is explained as follows.

The purpose of SNN is to mine TNN weights, thus we need to firstly extract the features in the weight matrix for making decisions for switching connections. Considering that the convolutional neural network (CNN) is a useful tool for extracting the features, we apply it to our problem. The CNN as the encoder subnetwork of our SNN is shown in Figure 4, which is composed of multiple blocks of convolution, ReLU activation, and max pooling. Based on the extracted features, we add a decoder subnetwork to obtain the values reflecting the importance of connections. In this part, we only upsample the widths of decoded feature maps, and the corresponding channel numbers do not change. The feature map obtained by downsampling is spliced with the corresponding upsampling feature map through row convolution.

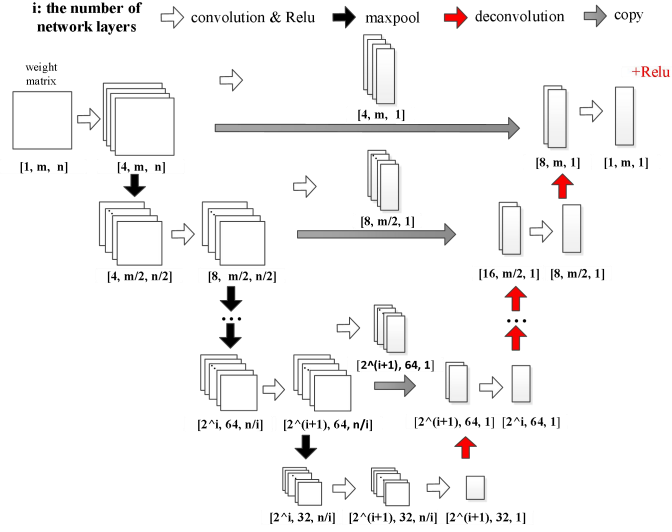

Figure 4: SNN structure

Noted that in the first strategy of switching TNN neurons, because the size of SNN output is not same as its input, the encoded features cannot be directly copied to the corresponding layer of the decoder. We introduce a row convolution to transform the encoded features to the same shape of decoded feature map. Furthermore, the widths of decoded feature maps are always 1 to determine the weighting degree of each neuron.

Finally, the SNN output values through ReLU activation function, where "0 "results in the pruning of connections, "<1" weakening, and ">1" strengenthening.

| Model | Method | Acc (%) | Params | FLOPs |
|---|---|---|---|---|
| LeNet-300-100 | Baseline [12] | 98.36 | 266K | 1.0× |
| | DDP [12] | 98.38 | 8.2K | 32.4× |
| | $L0_{hc}$ [24] $\lambda = 0.1/N$ | **98.60** | 69.3K | 3.8× |
| | $L0_{hc}$ [24] $\lambda$ sep. | 98.20 | 26.6K | 10.0× |
| | SNN neuron | 98.42 | 47.7K | 5.6× |
| | SNN connection | 98.43 | 42.5K | 6.3× |
| LeNet5-Caffe-20-50-800-500 | $L0_{hc}$ [24] $\lambda = 0.1/N$ | 99.10 | *43.4K | - |
| | $L0_{hc}$ [24] $\lambda$ sep. | 99.00 | *20.6K | |
| | SNN neuron | **99.27** | *61.7K | |
| | SNN connection | **99.31** | *57.1K | |

Table 1: Results on MNIST datasests. Params denotes the statistics of network parameters, and * represents that only sparse learning is performed on the fully connected layers.

| Model | Method | Acc (%) | Params | FLOPs |
|---|---|---|---|---|
| ResNet18 | N2N [40] | 91.97 | - | - |
| | SNN neuron | **93.44** | | |
| | SNN connection | **93.37** | | |
| ResNet34 | N2N [40] | 93.54 | - | - |
| | SNN neuron | **93.81** | | |

| Model | Method | Acc | Params | FLOPs |
|---|---|---|---|---|
| | SNN connection | **93.81** | | |
| VggNet16 | Baseline [17] | 92.52 | 38.9M | 1.0× |
| | NestedNet [41], L | 91.29 | 19.4M | 2.0× |
| | NestedNet [41], H | 92.40 | 38.9M | 1.0× |
| | DEN (ρ = 0.1) [11] | 92.31 | 18.5M | 2.4× |
| | SNN neuron | **92.94** | *25.4M | 1.5× |
| MobileNet_V1 | MobileNet_V1(×0.25) [42] | 86.3 | 0.5M | |
| | TD ρ = 0.95 [43] | 89.2 | - | |
| | Lottery Ticket (one-shot) [8] | 87.9 | - | - |
| | DNW(×0.225) [44] | 89.7 | - | |
| | SNN neuron | **91.0** | *0.48M | |
| | SNN connection | **90.4** | *0.47M | |

Table 2: Results on CIFAR10 datasests. Params denotes the statistics of network parameters, and * represents that only sparse learning is performed on the fully connected layers.

| Model | Method | Acc (%) | Params | FLOPs |
|---|---|---|---|---|
| ResNet18 | N2N [14] | 69.64 | | |
| | SNN neuron | **72.37** | - | - |
| | SNN connection | **72.38** | | |
| ResNet34 | N2N [14] | 70.11 | | |
| | SNN neuron | **72.98** | - | - |
| | SNN connection | **73.06** | | |
| VggNet16 | Baseline [17] | **69.64** | 38.9M | 1.0× |
| | NestedNet [41], L | 68.10 | 19.4M | 2.0× |
| | NestedNet [41], H | 69.01 | 38.9M | 1.0× |
| | DEN (ρ = 0.1) [11] | 68.87 | 18.9M | 1.7× |
| | SNN neuron | 68.29 | *28.7M | 1.3× |

Table 3: Results on CIFAR100 datasests. Params denotes the statistics of network parameters, and * represents that only sparse learning is performed on the fully connected layers.

## 4. Experiments

In this section, we demonstrate the effectiveness of our SNN-TNN learning framework for image classification. First, we explore the use of SNN on various datasets and model classifiers to train sparse neural networks without retraining or fine-tuning. Second, we compare SNN with other methods of learning network structures.

In all of the experiments, we use the stochastic gradient descent (SGD) as the optimizer with the batch size of 64 and automatic adjusting of learning rates. It starts with a learning rate of 0.1. For the random initialization, we use Kaiming normal distribution [36]. The experiments are conducted in PyTtorch platform [37].

### 4.1 MNIST CLASSIFICATION

The MNIST datasets contain 60,000 hand-written digits for training and 10,000 for testing. Each data point is of size 28 × 28 = 784 dimensions, and there are 10 classes of labels. LetNet-300-100 [41] used in this application has two hidden layers, with 300 and 100 neurons for each.

LetNet5-Caffe-800-500 has three convolutional layers and two fully connected layers with 800 and 500 neurons for each.

We report the learning results from our approach in Table 1, including the classification accuracy of trained neural networks and the parameters of the network. It shows that our method can effectively obtain the sparse fully connected layers that don't sacrifice the classification accuracy but even improve it. Especially for LetNet5-Caffe-800-500, our test accuracy is higher, according to the Ockham's Razor principle. These phenomena show that our SNN can choose a more reasonable structure on the task neural network.

## 4.2 CIFAR10 CLASSIFICATION

CIFAR10 [39] consists of 50,000 training images and 10,000 test images. There are 10 classes of labels. ResNet18 and ResNet34 have seventeen and thirty-three convolutional layers, respectively, and one fully connected layer. VggNet16 has thirteen convolutional layers and three fully connected layers and MobileNet_V1.

The testing results in this experiment are shown in Table 2. The results show that the classification accuracy on all three networks can be improved by using our approach. Especially for VggNet16, our method achieve better performance than Baseline, in the case where other methods result in slightly lower performance. Compared with ResNet, VggNet has more than three layers of complete connectivity. Therefore, our method may have better results when applied to the cases with more layers that can be processed by our approach.

## 4.3 CIFAR100 CLASSIFICATION

CIFAR100 used in this experiment consists of 50,000 images for training and 10,000 for testing. and there are 100 classes of labels. We use classical ResNet18, ResNet34 and VggNet16 to do the classification for CIFAR100.

Table 3 lists the accuracy and model parameter results achieved by our approach in this application. We can see that ResNet18 and ResNet34 still have good accuracy results, which are greatly improved compared with N2N. Why is the sparsity of ResNet18, ResNet34 and VggNet16 in this application less than that in CIFAR10 application? We think the reason is that CIFAR100 classification is more difficult than CIFAR10 classification, thus more complicated neural network is needed to solve it. More complicated neural networks will lead to bigger sparsity, and for the same network, more difficult tasks will result in less sparsity. This phenomenon proves that our approach is able to choose the appropriate structures adapting to the applications.

## 5. Conclusions

This paper has proposed a novel sparsity learning of neural networks and applied it to obtain sparse fully connected layers in classification networks. Our main idea is to introduce a switcher neural network (SNN) for utilizing the knowledge embedded in networks weights to switch the connections in the task neural network (TNN). The training of SNN is coupled with that of TNN. They are alternated with each other to produce optimal structure and parameters of TNN simultaneously. We test our method on various classical deep neural networks and commonly used datasets. A steady increase in accuracy along with the sparsity of network is observed by using our approach. Furthermore, more sparseness can be obtained for more complicated network and/or simpler applications. We can conclude that our approach has the ability to choose the appropriate sparse networks automatically.

In this work, only fully connected layers are considered for testing our approach. In its principal, the SNN can be applied to learn other network structures. In the next work, we want to extend and test the proposed approach for the learning of sparse convolutional layers.